\documentclass[authoryear]{elsarticle}
\usepackage[hidelinks]{hyperref}
\usepackage[dvipsnames,table]{xcolor}
\usepackage{amsmath}
\usepackage{amssymb}
\usepackage{csquotes}
\usepackage{etoolbox,siunitx}
\robustify\bfseries
\usepackage{commath}
\usepackage{booktabs}
\usepackage{multirow}
\usepackage[capitalise]{cleveref}
\usepackage{pgf} 
\usepackage{layouts} 
\usepackage{adjustbox}
\usepackage{mathtools}
\usepackage{enumitem}
\usepackage[printonlyused]{acronym}
\usepackage{tabularx}

\usepackage[ruled, vlined, linesnumbered]{algorithm2e}

\DeclareMathOperator*{\argmax}{arg\,max}
\DeclareMathOperator*{\argmin}{arg\,min}
\DeclareMathOperator{\ccr}{\textit{CCR}}
\DeclareMathOperator{\mae}{\textit{MAE}}
\DeclareMathOperator{\amae}{\textit{AMAE}}
\DeclareMathOperator{\mmae}{\textit{MMAE}}
\DeclareMathOperator{\gms}{\textit{GMS}}
\DeclareMathOperator{\ms}{\textit{MS}}
\DeclareMathOperator{\gmsp}{\textit{GMSp}}
\DeclareMathOperator{\msp}{\textit{MSp}}
\DeclareMathOperator{\auc}{\textit{AUC}}
\DeclareMathOperator{\wk}{\kappa}
\DeclareMathOperator{\spearman}{r_s}
\DeclareMathOperator{\kendall}{\tau_b}

\DeclareMathOperator{\Cov}{Cov}
\newcommand*{\xbf}{\mathbf{x}}
\newcommand*{\class}{\mathcal{C}}
\newcommand*{\neigh}{\mathcal{N}}

\newcommand{\Ioflupane}{Ioflupane ($^{123}$I)}
\newcommand{\ogosp}{OGO-SP}
\newcommand{\ogospbeta}{OGO-SP-$\beta$}
\newcommand{\maximize}{($\uparrow$)}
\newcommand{\minimize}{($\downarrow$)}
\newcommand{\rev}[1]{#1}
\newcommand*{\mrev}{}
\newcommand{\subscript}[1]{\ensuremath{_{\textrm{#1}}}}

\journal{Expert Systems with Applications}

\begin{document}
	
	\begin{frontmatter}
		\title{An ordinal CNN approach for the assessment of neurological damage in Parkinson's disease patients\tnoteref{fund}}
		\tnotetext[fund]{Funding: This work has been partially subsidised by the Spanish Ministry of Economy and Competitiveness (MINECO) [project reference TIN2017-85887-C2-1-P], the \enquote{Consejería de Econom\'ia, Conocimiento, Empresas y Universidad} of the \enquote{Junta de Andaluc\'ia} [project reference UCO-1261651] and FEDER funds of the European Union. Javier Barbero-G\'omez research has been subsidised by the FPI Predoctoral Program of the Spanish Ministry of Science, Innovation and Universities (MCIU) [grant reference PRE2018-085659]. V\'ictor-Manuel Vargas's research has been subsidised by the FPU Predoctoral Program of the MCIU, [grant reference FPU18/00358].}
		
		\author[ayrna]{Javier Barbero-G\'omez\corref{cor1}}
		\ead{jbarbero@uco.es}
		\author[ayrna]{Pedro-Antonio Guti\'errez}
		\ead{pagutierrez@uco.es}
		\author[ayrna]{V\'ictor-Manuel Vargas}
		\ead{vvargas@uco.es}
		\author[reinasofia]{Juan-Antonio Vallejo-Casas}
		\ead{jantonio.vallejo.sspa@juntadeandalucia.es}
		\author[ayrna]{C\'esar Herv\'as-Mart\'inez}
		\ead{chervas@uco.es}
		
		\address[ayrna]{Department of Computer Science and Numerical Analysis, University of C\'ordoba, Rabanales Campus, Albert Einstein building, 14014 C\'ordoba, Spain
				}
		\address[reinasofia]{UGC Medicina Nuclear, Hospital Universitario \enquote{Reina Sof\'ia}. University of C\'ordoba. IMIBIC. 14004 C\'ordoba, Spain.\newpage
		}
	
		\cortext[cor1]{Corresponding author}
		\begin{abstract}
			3D image scans are an assessment tool for neurological damage in \ac{PD} patients. This diagnosis process can be automatized to help medical staff through \acp{DSS}, and \acp{CNN} are good candidates, because they are effective when applied to spatial data. This paper proposes a 3D \ac{CNN} ordinal model for assessing the level or neurological damage in \ac{PD} patients. Given that \acp{CNN} need large datasets to achieve acceptable performance, a data augmentation method is adapted to work with spatial data. We consider the Ordinal Graph-based Oversampling via Shortest Paths (\ogosp{})\acused{OGO-SP} method, which applies a gamma probability distribution for inter-class data generation. A modification of \ac{OGO-SP} is proposed, the \ogospbeta{} algorithm, which applies the beta distribution for generating synthetic samples in the inter-class region, a better suited distribution when compared to gamma. The evaluation of the different methods is based on a novel 3D image dataset provided by the Hospital Universitario \enquote{Reina Sof\'ia} (C\'ordoba, Spain). We show how the ordinal methodology improves the performance with respect to the nominal one, and how \ogospbeta{} yields better performance than \ac{OGO-SP}.
		\end{abstract}
		
		
	\end{frontmatter}

	\acrodef{PD}{Parkinson's disease}
	\acrodef{CAD}{Computer-Aided Diagnosis}
	\acrodef{DSS}{Decision Support System}
	\acrodef{CCR}[$\ccr$]{Correct Classification Rate}
	\acrodef{MAE}[$\mae$]{Mean Absolute Error}
	\acrodef{WK}[$\wk$]{Weighted Cohen's Kappa}
	\acrodef{KR}[$\kendall$]{Kendall's rank correlation coefficient}
	\acrodef{SR}[$\spearman$]{Spearman's rank correlation coefficient}
	\acrodef{UPDRS}{Unified Parkinson's Disease Rating Scale}
	\acrodef{MRI}{Magnetic Resonance Imaging}
	\acrodef{SPECT}{Single-Photon Emission Computerized Tomography}
	\acrodef{PET}{Positron Emission Tomography}
	\acrodef{PPMI}{Parkinson's Progression Markers Initiative}
	\acrodef{LCC}{LRRK2 Cohort Consortium}
	\acrodef{ROI}{Region of Interest}
	\acrodefplural{ROI}[ROIs]{Regions of Interest}
	\acrodef{ML}{Machine Learning}
	\acrodef{POM}{Proportional Odds Model}
	\acrodef{CLM}{Cumulative Link Model}
	\acrodef{CNN}{Convolutional Neural Network}
	\acrodef{GAN}{Generative Adversarial Network}
	\acrodef{LReLU}{Leaky Rectified Linear Unit}
	\acrodef{OBD}{Ordinal Binary Decomposition}
	\acrodef{ECOC}{Error Correcting Code}
	\acrodefplural{ECOC}[ECOCs]{Error Correcting Codes}
	\acrodef{SMOTE}{Synthetic Minority Oversampling Technique}
	\acrodef{OGO}{Ordinal Graph-based Oversampling}
	\acrodef{OGO-SP}{OGO via Shortest Paths}
	\acresetall
	
	
	\section{Introduction}
	
	In the medical field, plenty of data is obtained from patients in order to give an accurate diagnosis and decide on a beneficial treatment. From simple scalar to fully structured spatiotemporal data, like electrocardiograms, projectional radiographs, and computed tomography scans, all of these sources aid medical staff look for abnormalities and signs of degradation or damage. Nonetheless, these examinations are time-consuming and costly for hospitals and other medical administrations. A great deal of expertise is needed in order to interpret all this data into useful information.
	
	\subsection{Machine Learning}
	\ac{ML} techniques can automatize some of these tasks by leveraging on existing data. They can serve as the core to a \ac{DSS} in order to help medical staff make a better judgment or give an additional opinion.
	
	In the case of structured data, \acp{CNN} have shown excellent performance in tasks such as classification, segmentation or regression. These models rely on large amounts of labeled data in order to learn. Examples of such tasks can be diagnosis (classification) \citep{el-dahshanComputeraidedDiagnosisHuman2014,mazaheriHeartArrhythmiaDiagnosis2020}, automatic identification of anatomical regions (segmentation) \citep{pengLCPNetLocalContextperception2020} or landmark location (regression) \citep{payerIntegratingSpatialConfiguration2019a}.
	
	\subsection{Parkinson's disease}
	
	One of many medical applications of \ac{ML} explored in the literature is Parkinson's disease (PD)\acused{PD} diagnosis. \ac{PD} is a degenerative nervous system disorder affecting the brain whose symptoms are primarily motor-related: shaking, gait disturbances, slowness and difficulty walking. Other symptoms are related to sleeping, emotional or sensory problems. The cost on society of this disease grows as the symptoms worsen, as the greatest component of cost is in patient care and nursing home costs. Just in the UK, the total cost has been estimated to be between \SI{445}[\pounds]{} million and \SI{3.3}[\pounds]{} billion annually \citep{findleyEconomicImpactParkinson2007}.
	
	Assessing the severity of the neurological damage of \ac{PD} patients is a crucial part for the correct treatment, as an unnecessarily high dose of levodopa (the most common medication for \ac{PD}) may worsen symptoms in the long-term  \citep{tomlinsonSystematicReviewLevodopa2010}. To evaluate this, doctors rely both on observations of motor capabilities \citep{marinoMagneticResonanceImaging2012} as well as imaging techniques like \ac{MRI} \citep{marinoMagneticResonanceImaging2012} and nuclear tomography like \ac{SPECT} or \ac{PET} \citep{arbizuFunctionalNeuroimagingDiagnosis2014}. \Ioflupane{} (known commercially as \enquote{DaTscan}) is a neuro-imaging drug often used to evaluate the dopaminergic activity in the nigrostriatal dopaminergic pathway when the disease may be in the early stages \citep{darcourtEANMProcedureGuidelines2010}. It is injected into the patient's bloodstream before taking a \ac{SPECT} image, so that the brain's dopaminergic activity can be inspected visually. The evaluation of this damage may require a great deal of experience.
	
	In recent years there has been a growing interest on the application of \ac{ML} techniques to this kind of images, which require no previous assumptions on the regions of interest or relevant areas for the given task. Instead, these methods are able to discover for themselves where and what to look for in images, based solely on data previously labeled by doctors. Popular methods tackle binary (\enquote{healthy control} or \enquote{disease}) \citep{viniciusdossantosferreiraConvolutionalNeuralNetwork2018} or nominal (\enquote{healthy control}, \enquote{disease A}, \enquote{disease B}, ...) classification \citep{akyolStackingEnsembleBased2020}.
	
	Some datasets related to \ac{PD} are available online for research use, such as the \ac{PPMI}\footnote{\url{http://www.ppmi-info.org/}} or the \ac{LCC}\footnote{\url{https://www.neurodegenerationresearch.eu/cohort/lrrk2-cohort-consortium/}}.
	
	\subsection{Ordinal classification}
	
	In the last decade, the concept of \enquote{ordinal classification} (sometimes referred as \enquote{ordinal regression}) has been popularized as a way to exploit extra information in a classification problem where a natural ordering of the classes is present \citep{gutierrezOrdinalRegressionMethods2016,ben-davidComparisonClassificationAccuracy2008}. This approach has been proven to outperform the classic nominal perspective in medical applications such as melanoma diagnosis \citep{sanchez-monederoPartialOrderLabel2018} and liver transplantation \citep{dorado-morenoDynamicallyWeightedEvolutionary2017}. Up until now, research into this methodology applied to \ac{PD} diagnosis has not been addressed, because existing scales, such as the Hoehn and Yahr scale \citep{hoehnParkinsonismOnsetProgression1967} or the \ac{UPDRS}, require subjective evaluation from clinicians of tremor, rigidity or movement \citep{ramakerSystematicEvaluationRating2002} and are difficult to quantify in a single measure.
	
	A complete taxonomy of ordinal classification methods can be found in \cite{gutierrezOrdinalRegressionMethods2016}. The most naive methods perform a simple regression using the class labels and then round the values when predicting \citep{kramerPredictionOrdinalClasses2010} or simply apply a label distance cost penalty to a classical nominal classification method \citep{kotsiantisCostSensitiveTechnique2004}. These basic approaches lack an understanding of the underlying label distance, as different misclassifications may represent a different error cost.
	
	Threshold Models are another popular approach for this task. An underlying latent continuous variable is assumed to exist, from which the different ranks arise by assigning certain thresholds. Thus, in this framework, both the value of the latent variable and the thresholds need to be learned from the data. Some approaches, like the classical \ac{POM} \citep{mccullaghRegressionModelsOrdinal1980a} or the more recent \texttt{gologit} model \citep{williamsGeneralizedOrderedLogit2006} fall into the \acp{CLM} framework, a probabilistic method for predicting probabilities of groups of contiguous categories, taking the ordinal scale into account.
	
	Other ordinal approaches consist on decomposing the ordinal problem into a set of binary problems (called \ac{OBD}). Sometimes these decompositions are solved by a set of different models, like in the cascade linear utility model \citep{wuPracticalSVMbasedAlgorithm2003}. In other cases they are modeled by several outputs of the same underlying model \citep{jianlinchengNeuralNetworkApproach2008a}. All \acp{OBD} present the same challenge: combining the results of all decompositions into a single final classification. The simplest approaches assign the first class to reach a certain threshold \citep{wuPracticalSVMbasedAlgorithm2003}, but this can lead to an imbalance as the last classes are more difficult to select. \acp{ECOC} are better suited to this task, as this approach considers all outputs equally in the final decision \citep{allweinReducingMulticlassBinary2001}.
	
	The available \ac{PD} datasets mentioned previously only provide label information of binary, nominal or continuous nature. Our study presents a novel dataset, collected by the Clinical Management Unit of Nuclear Medicine of the Hospital Universitario \enquote{Reina Sof\'\i{}a} (C\'ordoba, Spain), containing \ac{SPECT} \Ioflupane{} images of \ac{PD} patients classified in 4 distinct ordinal labels, according to their stage of presynaptic dopamine binding damage  (\enquote{no alteration}, \enquote{slight alteration}, \enquote{more advanced alteration} and \enquote{severe alteration}). Ordinal methods are perfect candidates to tackle the task of predicting these labels, which serve as a better indicator for the medical decision process.
	
	\subsection{Data augmentation}
	
	Classification tasks may suffer from unbalanced data, especially in the medical field, as healthy patients are much more common than sick patients. Also, data gathering and proper labeling is expensive and time consuming. In this situation, data augmentation techniques are required in order to boost the performance of \ac{ML} models.
	
	The most basic strategy for augmenting spatial data, like medical images, is image translation, rotation, flipping and cropping \citep{perezEffectivenessDataAugmentation2017}. Several of these can be used depending on the specific task to learn. For example, object detection tasks, such as anomaly or lesion detection, can be accelerated by using cropped \acp{ROI} as samples \citep{reyHybridCADSystem2020}.
	
	Classic techniques like Synthetic Minority Oversampling (\acs{SMOTE})\acused{SMOTE} \citep{chawlaSMOTESyntheticMinority2002,riveraPrioriSyntheticOversampling2016} perform well on low dimensional data. Some techniques, such as Autoencoders \citep{hintonAutoencodersMinimumDescription1993} or \acp{GAN} \citep{goodfellowGenerativeAdversarialNetworks2014} are able to leverage convolutional operations, which improve the performance and efficiency over spatial data. Those techniques, however, require a vast amount of training examples in order to avoid pitfalls like mode collapse.
	
	In order to meet this data volume requirement, more sophisticated data augmentation methods are being applied to medical data as of recently. As an example, \cite{salazarGenerativeAdversarialNetworks2021} combine \acp{GAN} with Markov Random Field models to augment 3D functional \ac{MRI} multi-subject data and enhance nominal classification performance.
	
	There also exist data augmentation methodologies that make use of ordinal information in the class labels to improve the synthetic data generation process. A family of such methods, presented by  \cite{perez-ortizGraphBasedApproachesOverSampling2015}, are the \ac{OGO} methods. These consist on computing a graph estimating the latent manifold structure in the data by exploiting ordinal information in the labels, and then using the edges on that graph to generate samples, similar to \ac{SMOTE}.
	
	Classic proven techniques such as \ac{SMOTE} and \ac{OGO} can be adapted to work on spatial data: a \ac{CNN} can first be trained so that a projection from high-dimensional data is learned, and then a traditional data augmentation method can be applied to the resulting low-dimensional data.
	
	\subsection{Limitations of OGO}
	
	\cite{perez-ortizGraphBasedApproachesOverSampling2015} propose the use of a gamma distribution for the generation of samples in the inter-class space of the latent manifold. This distribution is skewed towards the part of the graph adjacent to the class to be augmented but permits the inclusion of features in the frontier of both classes.
	
	This, however, presents some limitations. The gamma distribution is not suited for the generation of values in a closed domain (in this case, $\delta \in [0, 1]$), as its original domain is in the interval $[0,\infty)$. This is not only a theoretical overlook, but it also hinders the ability to tune its parameters in a meaningful way, subject to the dataset to which it is applied.
	
	In this paper, an alternative to the gamma distribution is proposed: using the better suited beta distribution, which is bounded in the $[0, 1]$ interval and has easier-to-tune parameters, enabling a higher degree of flexibility which can help achieve better performance.
	
	\subsection{Goals}
	
	To the best of our knowledge, there are no previous works using \ac{ML} methods for the assessment of the severity of brain damage from a patient's brain \ac{SPECT} 3D image. These methods could help doctors in the diagnosis and treatment of \ac{PD} and other parkinsonisms through \acp{DSS} and contribute towards the relief of the public health cost of this disease.
	
	Thus, the goals of the present work are:
	
	\begin{enumerate}
		\item Exploring the potential classification performance improvement in using ordinal label information.
		\item Adapting the use of classical data augmentation and class balancing techniques to spatial three-dimensional data.
		\item Analyzing the developed methodologies in points 1 and 2 using a novel and extensive database of \ac{SPECT} images from Hospital \enquote{Reina Sof\'\i{}a} (C\'ordoba, Spain).
		\item Studying a potential improvement to the data augmentation methodology presented by \cite{perez-ortizGraphBasedApproachesOverSampling2015} by applying a better suited probability distribution for generating synthetic samples in the class frontiers.
	\end{enumerate}
	
	The rest of this paper is organized as follows: In \cref{sec:materials-methods} we describe the data to be used and we propose a fully 3D Deep CNN model for the evaluation of presynaptic deficit in \ac{SPECT} \Ioflupane{} images. We design two versions of the same base model, one nominal and the other ordinal, differing on the output layer shape and the activation function as well as the loss function. We propose a novel approach for data augmentation using the beta probability distribution with efficiently estimated parameters. In \cref{sec:experimentation}, we describe the design of the experiments and metrics for the evaluation of the proposed methods. We show that our proposal performs better than the nominal methodology as well as the previous ordinal method. Finally, in \cref{sec:conclusions}, we discuss the results and propose future work.
	
	\section{Materials and methods} \label{sec:materials-methods}
	
	In this section, we present the dataset used for the experimentation, as well as the architecture of the proposed models and the novel data augmentation proposal.
	
	These models have been designed to tackle the task of assessing the alteration of dopaminergic activity in the brain of \ac{PD} patients by examining 3D \ac{SPECT} scans of the brain.
	
	\subsection{Data description} \label{sec:data-description}
	
	\rev{The dataset consists of 508 3D images provided by the Clinical Management Unit of Nuclear Medicine of the Hospital Universitario \enquote{Reina Sof\'\i{}a} (C\'ordoba, Spain). They are obtained by first administering the patients with \Ioflupane{}, a radiopharmaceutical which binds to the presynaptic dopamine transporters in the brain. Some time later, a \ac{SPECT} scan is performed, so as to inspect the dopaminergic activity in the nigrostriatal dopaminergic pathway, which is one of the neuropathological characteristics of \ac{PD}.}
		
	\rev{Of these images, 314 (\SI{61.8}{\percent}) are of healthy patients (class 0), 42 (\SI{8.3}{\percent}) show slight alteration (class 1), 52 (\SI{10.2}{\percent}) show more advanced alteration (class 2) and 100 (\SI{19.7}{\percent}) show severe alteration (class 3). It is common for medical diagnosis datasets like this one to have a severe imbalance problem. In our case, more than \SI{60}{\percent} of samples are of healthy patients, and less than \SI{10}{\percent} belong to class 1.}
	
	\rev{The doctors' diagnosis is attached as a class label to each of the images. Because of the gradual nature of these classes, the task of recognizing which of the four categories an image belongs to can be posed as an ordinal classification problem and, thus, specific techniques can be employed to exploit the order information.}
	
	Automatic linear image registration has been performed on all images using the FMRIB's Linear Image Registration Tool (FLIRT) from the FMRIB Software Library (FSL) \citep{smithAdvancesFunctionalStructural2004} considering the T2 version of the MNI152 2mm standard space \ac{SPECT} template \citep{evansBrainTemplatesAtlases2012}. Thus, all images have a final resolution of $91 \times 109 \times 91$ voxels. Also, during training, the symmetrical nature of the images is exploited: with \SI{50}{\percent} probability, the images are flipped on the frontal axis (left to right) each time they are used.

	\subsection{Global architecture}
	
	The overall architecture of the \ac{CNN} model considered in this paper consists on convolutional blocks of repeating layers reducing the size of the image while increasing the number of feature maps. Afterwards, the output of the convolutional part of the network is fed into a fully connected layer of neurons before computing the output decision of the model.
	
	Each convolutional block consists of a 3D convolutional layer followed by a batch normalization layer. The kernel size and the stride of the convolution are parameters to be cross-validated during the training process. The output of each block is then fed into a \ac{LReLU} activation function, which has been proven to have good convergence properties such as scale-invariance and 1-Lipschitz continuity \citep{suzukiFastGeneralizationError2018}.
	
	The low resolution feature maps which are the output of the convolutional blocks are then the input of a densely connected neuron layer (the number of neurons of this layer is also cross-validated during training) using, again, \ac{LReLU} as the activation function. A final output layer computes the final classification given by the model. In the training phase, the model weights are updated using the Adam optimization algorithm \citep{kingmaAdamMethodStochastic2017} so that the outputs align with the annotated labels.
	
\begin{figure}
		\includegraphics[width=\linewidth]{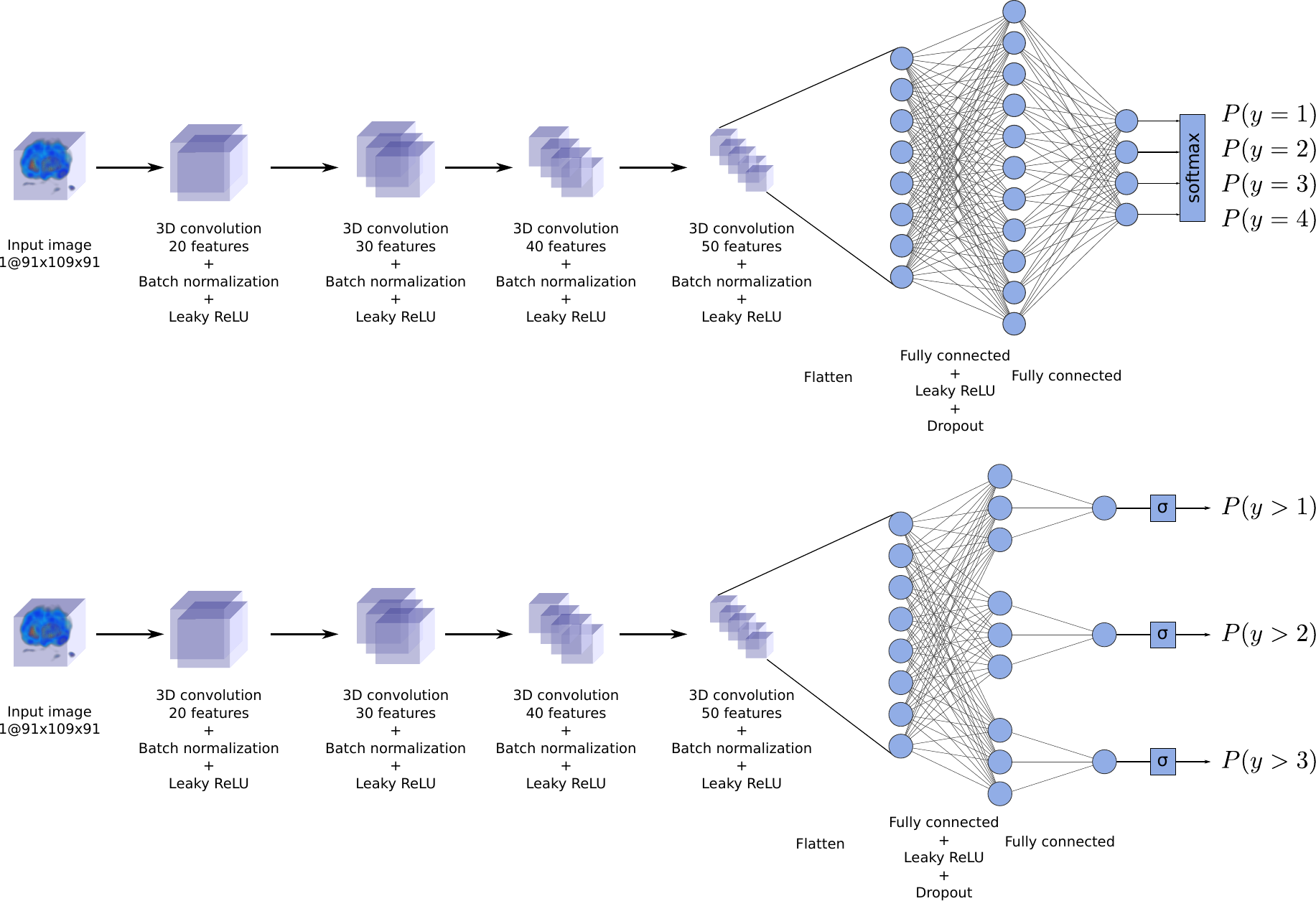}
		\caption{The two network architectures: nominal (above) and ordinal (below).}
		\label{fig:architecture}
	\end{figure}

	We test two different architectures: a classic architecture based on nominal classification and an ordinal architecture, considering the ordering of the class labels. In \cref{fig:architecture}, it can be noted that both architectures share the same structure for their convolutional part, but they are different in the way the final output is computed. Moreover, two different class balancing methods are applied in each case, as explained in \cref{sec:class-balancing}.
	
	\subsection{Classic nominal classification architecture}
	
	As is the case with classic \ac{CNN} architectures, the output of the convolutional part of the network is then fed to a single fully connected layer. This is then again fully connected with the output layer, which has as many outputs as classes to decide. Then, a softmax activation function maps the output of the network into a set of probabilities $o_q$ of belonging to each class $\class_q$: $o_q = P(y=\class_q)$.
	
	The optimizing criterion is the categorical cross-entropy loss, described as:
	\begin{align}
		\ell(x_i) &= -\sum_{q=1}^Q 1\{y_i = \class_q\} \log(P(y_i = \class_q \, | \, \xbf_i)),
	\end{align}
	where $Q$ is the number of classes, $1\{y_i = \class_q\}$ is the indicator function that is equal to 1 when $y_i = \class_q$ and 0 otherwise and $P(y = \class_q \, | \, \xbf_i)$ is the predicted probability of $\xbf_i$ belonging to class $\class_q$.
	
	To evaluate the effectiveness of the trained model, a new unseen given sample $\xbf$ is classified as belonging to the class with maximum predicted probability
	\begin{equation*}
		\hat{y} = \argmax\limits_{1 \leq q \leq Q} P(y = \class_q \, | \, \xbf).
	\end{equation*}
	
	\subsection{Proposed ordinal classification architecture}
	
	When the classes are ordered, instead of dealing with the complete problem as previously mentioned, \ac{OBD} can be applied: the problem is decomposed into $Q-1$ binary decision problems. Each problem $q$ consists on deciding if $y \succ \class_q$ conditioned to $\xbf$ ($1 \leq q < Q$).
	
	This would normally require $Q-1$ different models, each solving one of these binary problems.
	\rev{This approach, originally presented by \cite{frankSimpleApproachOrdinal2001}, would imply to first compute every probability $p_q = P(y = \class_q)$ based on the obtained models and then select the highest probability class. The individual probabilities are computed as a function of the cumulative probabilities, $P(y \succ \class_q)$, estimated by the binary models:
	\begin{align*}
		p_1 &= P(y = \class_1) = 1 - P(y \succ \class_1) \\
		p_q &= P(y = \class_q) = P(y \succ \class_{q-1}) - P(y \succ \class_{q}) \; \forall \, 1 < q < Q\\
		p_Q &= P(y = \class_1) = P(y \succ \class_{Q-1}).
	\end{align*}}
	\rev{However, different problems are associated to this approach: because the outputs of the different decompositions are not combined in the same training process, the basic probability assumptions (that is, $P(y \succ \class_{q}) \geq P(y \succ \class_{q+1})$, $p_q \geq 0$ and $\sum_{q} p_q = 1$) are not necessarily satisfied, which can lead to inconsistencies. Moreover, when computing the individual probabilities, at most only two of the model outputs are considered, instead of all of them. Finally, when obtaining the decomposition,} the ratio of positive to negative samples becomes very unbalanced in the extreme classes. In the case of an already unbalanced dataset, this procedure \rev{can become} unrealistic.
	
	To circumvent this limitation, a compromise is proposed: a single convolutional model can be trained simultaneously to then be fed into multiple fully connected blocks, each one solving an individual binary classification subproblem. This way, the imbalance of the training data can be less acute, and training can be done in parallel. The output of each of the $Q-1$ fully connected blocks has a sigmoid activation function representing the probability $o_k = P(y \succ \class_k \, | \, \xbf) \in (0, 1)$.
	
	\rev{Moreover, for obtaining the final probabilities, we use a more stable option based on the decision function of the \ac{ECOC} framework \citep{allweinReducingMulticlassBinary2001}. The correct ideal output code for each class is considered as the coordinates of a vertex of a hypercube in $Q-1$ dimensions, e.g. for a 4 class ordinal problem classes $\class_1$, $\class_2$, $\class_3$ and $\class_4$ would be associated to the codes $( 0,0,0 )$, $( 1,0,0 )$, $( 1,1,0 )$ and $( 1,1,1 )$, respectively. This way, all the model outputs are considered for classification. To decide which class a sample $\xbf$ belongs to, the class with the nearest code according to some distance $d$ is selected:
	\begin{equation*}
		\hat{y} = \argmin\limits_{1 \leq q \leq Q} \; d(\mathbf{o}, \mathbf{c}_q),
	\end{equation*}
	where $\mathbf{o} = ( o_1, o_2, ..., o_{Q-1} )$ is the vector of output values, and $\mathbf{c}_q$ is the code vector associated with class $\class_q$. It can be noted that the aforementioned probability assumptions do not need to hold in order to assign a class in this framework.}
	
	\rev{Instead of individually training the different binary subproblems (as proposed for ECOC) and} to be consistent with the decision criterion, the \rev{global} optimization criterion of the network is set to be the mean squared error ($MSE$) loss:
	\begin{equation*}
		\ell(\xbf_i) = \frac{1}{Q-1} \sum_{k=1}^{Q-1} (1\{y_i \succ \class_k\} - P(y_i \succ \class_k \, | \, \xbf_i))^2.
	\end{equation*}

	While classifying new samples, the $L_2$ norm is used as the distance metric $d$, as it aligns with the $MSE$ optimization criterion:
	\begin{equation*}
		\hat{y} = \argmin\limits_{1 \leq q \leq Q} \; \norm{\mathbf{o} - \mathbf{c}_q}_2.
	\end{equation*}
	
	\subsection{Class balancing} \label{sec:class-balancing}
	
	For problems of imbalanced nature, such as medical diagnosis, where samples of specific classes are a minority compared to the rest, some considerations need to be made in the training process to avoid biases in the final model that can undermine its generalization capabilities.
	
	Different techniques such as class balancing can help with this problem. These generally consist on sampling or generating techniques that balance the ratio of training samples of each class presented to the classifier during training.
	
	A popular approach is \ac{SMOTE}: in order to generate new samples of a minority class, several existing samples of said class from the training set are randomly selected and a random weighted sum of their features is used as a new synthetic sample of that class. \ac{SMOTE} can work well when the number of features is not too large and all features are continuous.
	
	In order to account for the ordinal information in the sample synthesis process, \ac{OGO} techniques based on the idea of SMOTE have been previously proposed, such as \ac{OGO} via shortest paths using a probability function for the inter-class edges (abbreviated to \ogosp{})\citep{perez-ortizGraphBasedApproachesOverSampling2015}. This algorithm defines a way to construct a graph which captures the neighbouring relations between samples of the dataset, considering the ordering information provided by the class labels.
	
	\subsection{\ogosp{} algorithm} \label{sec:ogo-sp}
	
	Consider a dataset $D = \{(\xbf_1, y_1), (\xbf_2, y_2), ..., (\xbf_N, y_N)\}$, where each $\xbf_i \in \mathbb{R}^d$ is a training sample and $y_i \in \{\class_1, \class_2, ..., \class_Q\}$ its corresponding class label, which follows a ranking structure ($\forall i < j, \; \class_i \prec \class_j$).
	
	From this, an $N$-vertex undirected graph $G = (V, E)$ will be constructed, where $V$ corresponds to the vertices representing the $N$ training samples and $E \subseteq V^2$ corresponds to the edges representing the neighbouring relations between samples:
	\begin{align*} 
		V &= \{v_1, v_2, ..., v_N\} = \{\xbf_1, \xbf_2, ..., \xbf_N\}, \\
		E &= \{e_{i,j}\} = \{(v_i, v_j)\} = \{(\xbf_i, \xbf_j)\}, \; 1 \leq i < j \leq N.
	\end{align*}
	
	If $q$ is the index of the class to be over-sampled, a graph $G'_q$ of this form will be constructed based on three subgraphs:
	\begin{align*}
		G_{q-1, q} &= (V_{q-1, q}, E_{q-1, q}),\\
		G_{q,q} &= (V_{q,q}, E_{q,q}),\\
		G_{q,q+1} &= (V_{q,q+1}, E_{q,q+1}),\\
		G'_q &= (V'_q, E'_q) = (V_{q-1, q} \cup V_{q,q} \cup V_{q,q+1}, E_{q-1, q} \cup E_{q,q} \cup E_{q,q+1}).
	\end{align*}

	The edges of graph $G_{q-1, q}$ are determined by the intersection of two different sets obtained from a neighbourhood analysis based on the distance relation $d$:
	\begin{align*}
		E_{q-1, q} &= \neigh_d(X_{q-1}, X_q, k) \cap \neigh_d(X_{q}, X_{q-1}, k), \\
		\neigh_d(X_1, X_2, k) &= \{e_{i,j} \; | \; (\xbf_i \in X_1) \land (\xbf_j \in X_2) \land (\xbf_j \in nn_d(\xbf_i, X_2, k))\},
	\end{align*}
	where $X_c$ is the subset of all samples with label $y = \class_c$, $\neigh_d(X_1, X_2, k)$ is the $k$-neighbourhood of $X_1$ with respect to $X_2$ according to some distance $d$ and $nn_d(\xbf, X, k)$ is the set containing the $k$ nearest neighbours of $\xbf$ from set $X$. The vertices $V_{q-1, q}$ are all those appearing on $E_{q-1, q}$:
	\begin{equation*}
		V_{q-1, q} = \{\xbf_i \; | \; ((\xbf_i, \xbf) \in E_{q-1, q}) \lor ((\xbf, \xbf_i) \in E_{q-1, q})\}.
	\end{equation*}

	Using only the intersection of both neighbourhoods ensures that only the connecting regions of each class are considered. Parameter $k$ will control how broad is the region to consider.
	
	Graph $G_{q, q+1}$ is defined in an analogous way:
	\begin{align*}
		E_{q, q+1} &= \neigh_d(X_{q}, X_{q+1}, k) \cap \neigh_d(X_{q+1}, X_{q}, k), \\
		V_{q, q+1} &= \{\xbf_i \; | \; ((\xbf_i, \xbf) \in E_{q, q+1}) \lor ((\xbf, \xbf_i) \in E_{q, q+1})\}.
	\end{align*}

	Finally, $G_q$ is simply defined as:
	\begin{align*}
		E_{q,q} &= \neigh_d(X_q, X_q, k), \\
		V_{q, q} &= \{\xbf_i \; | \; ((\xbf_i, \xbf) \in E_{q, q}) \lor ((\xbf, \xbf_i) \in E_{q, q})\}.
	\end{align*}

	For the case of the extreme classes ($\class_1$ and $\class_Q$), one of $G_{q-1, q}$ or $G_{q, q+1}$ may be empty and only the connecting region to the one adjacent class will be considered.
	
	Based on the ordinal classification hypothesis that the distance to adjacent classes is lower than the distance to non-adjacent classes, the final graph $G_q = (V_q, E_q)$ will be constructed based on the previously constructed $G'_q$. Ideally, a distance-based manifold exists in the class label, such that $X_q$ lies in the space between $X_{q-1}$ and $X_{q+1}$. In reality, some outliers may exist in $X_q$ that are not desirable in the over-sampling procedure. In order to identify the key samples which lie between the adjacent classes, the shortest paths between the samples of $X_{q-1}$ and $X_{q+1}$ are used to decide the edges present in the final $G_q$.
	
	A path between two vertices $v_1$ and $v_z$ of the graph is defined as the sequence $P = (v_1, v_2, ..., v_z) \in V^z$ such that $e_{i,i+1} = (v_i, v_{i+1}) \in E$. If a cost function $f: E \rightarrow \mathbb{R}$ assigning a cost to every edge is defined, the shortest path $P_{1,z}$ is that which minimizes the total sum of the costs of the edges $\sum_{i=1}^{z-1} f(e_{i,i+1})$. In our implementation, the cost function selected is the same as distance $d$ used for $\neigh_d$, which is the $L_2$ norm or euclidean distance:
	
	\begin{equation*}
		f(e_{i,j}) = d(\xbf_i, \xbf_j) = \norm{\xbf_i - \xbf_j}_2.
	\end{equation*}
	
	In order to find those patterns in $X_q$ lying in the latent ordinal manifold, all the shortest paths between all the vertices in $V_{q-1,q}$ and all in $V_{q,q+1}$ will be computed using Dijkstra's algorithm \citep{dijkstraNoteTwoProblems1959}, and only the edges contained in one or more of these paths will be included in $E_q$:
	\begin{align*}
		E_q = \{e_{i,j} \; | \; &\exists a \in V_{q-1,q}, b \in V_{q,q+1} \; \\
		                  &\text{s.t.} \; (v_i, v_j \in P_{a,b}) \lor (v_j, v_i \in P_{b,a}) \},
	\end{align*}
	\begin{equation*}
		V_q = \{ v_i \; | \; e_{i,j} \in E_q \}.
	\end{equation*}
	Note that, if $q$ is any of the extreme classes, $V_{q,q}$ will have to be used instead of $V_{q-1,q}$ or $V_{q,q+1}$, depending on the case.
	
\begin{figure}
		\input{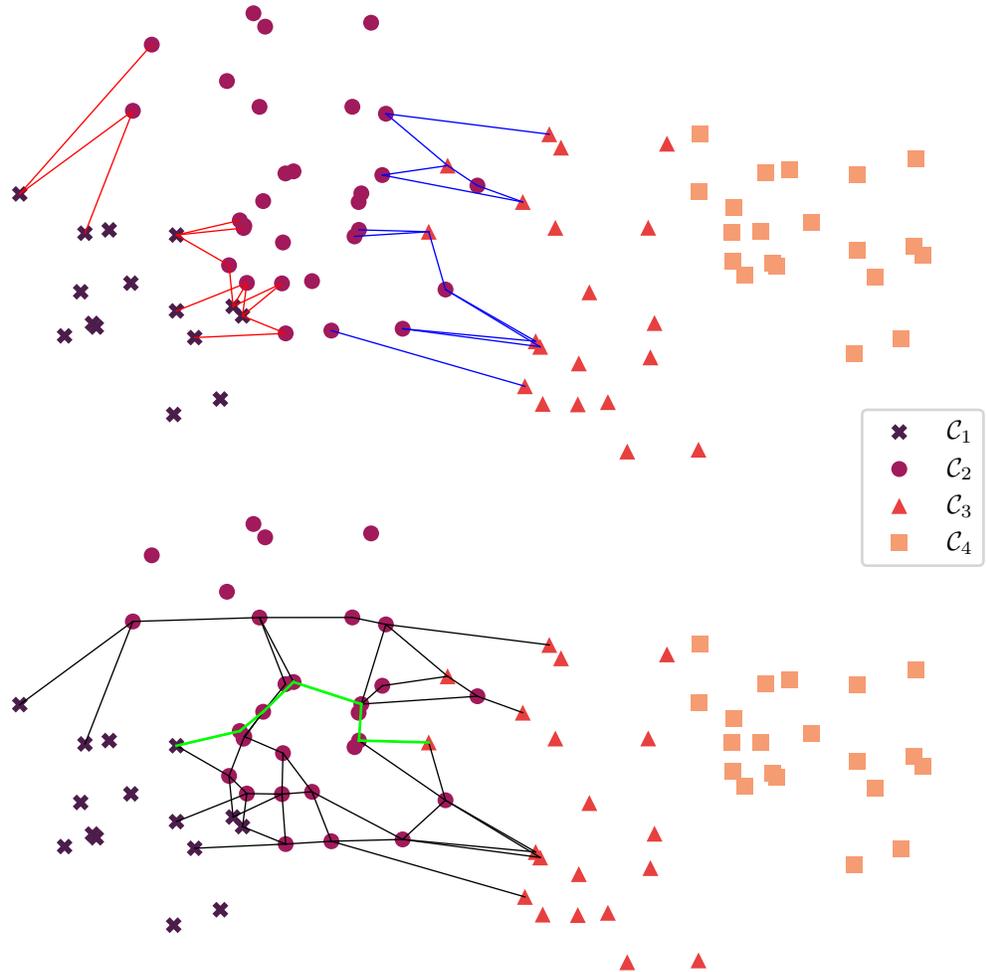}
		\caption{Example of the \ogosp{} graph construction procedure. The markers represent samples of the dataset. The graph $(V_2, E_2)$ corresponding to $\class_2$ is constructed. The top diagram shows $E_{1,2}$ (in red) and $E_{2,3}$ (in blue). The bottom diagram shows the shortest path between two vertices in $V_{1,2}$ and $V_{2,3}$ (in green) and the edges of the final constructed graph $E_2$ (in black).}
		\label{fig:graph-example}
	\end{figure}

	An example of the computed graph $(V_q, E_q)$ is shown in \cref{fig:graph-example}.
	
	Finally, new synthetic samples can be generated from $G_q$: in order to generate sample $(\xbf', \class_q)$, a random edge $e = (\xbf_i, \xbf_j) \in E_q$ is selected so that $\xbf'$ lies in the segment between $\xbf_i$ and $\xbf_j$:
	\begin{equation*}
		\xbf' = (1-\delta) \xbf_i + \delta \xbf_j,
	\end{equation*}
	where $\delta$ is a random variable in the range $[0, 1]$. The distribution from where $\delta$ is sampled will depend on the selected edge $e$:
	\begin{itemize}
		\item If both $y_i = \class_q$ and $y_j = \class_q$ (i.e. $e$ is an intra-class edge), then $\delta$ is sampled from a uniform  distribution $U(0, 1)$ just like \ac{SMOTE}.
		\item If $y_i = \class_q$ but $y_j \neq \class_q$ (i.e. $e$ is an inter-class edge), then $\delta$ is sampled from an asymmetrical distribution so that the new synthetic sample favours the augmented class but is able to capture the class transition phase. In the original \ogosp{} paper \citep{perez-ortizGraphBasedApproachesOverSampling2015}, $\delta \sim \text{Gamma}(k=2, \theta=0.15)$. While this has the previously mentioned properties, the gamma distribution is not bounded, $\delta \in [0, \infty)$, which means that $P(\delta > 1) > 0$, and there is a risk that $\xbf'$ lies in a different region of the manifold than $\class_q$ and is therefore incorrectly labelled.
	\end{itemize}

	In order to overcome this problem of the original algorithm, we propose the following  modification: instead of weighting on a parameter $\delta \sim \text{Gamma}(k, \theta)$, the better suited beta distribution is used $\delta \sim \text{Beta}(\alpha, \beta)$, where the parameters $\alpha>0$ and $\beta>0$ control the shape of the distribution \citep{keepingBetaDistribution1995}. \rev{We hypothesize that the beta distribution is better suited than the gamma distribution for this purpose, because it is bounded in the interval $[0,1]$, it has lower variance and its parametrization allows the probability density to be skewed towards a specific extreme, depending on the values of $\alpha$ and $\beta$.}
	
\begin{figure}
		\begin{center}
			\input{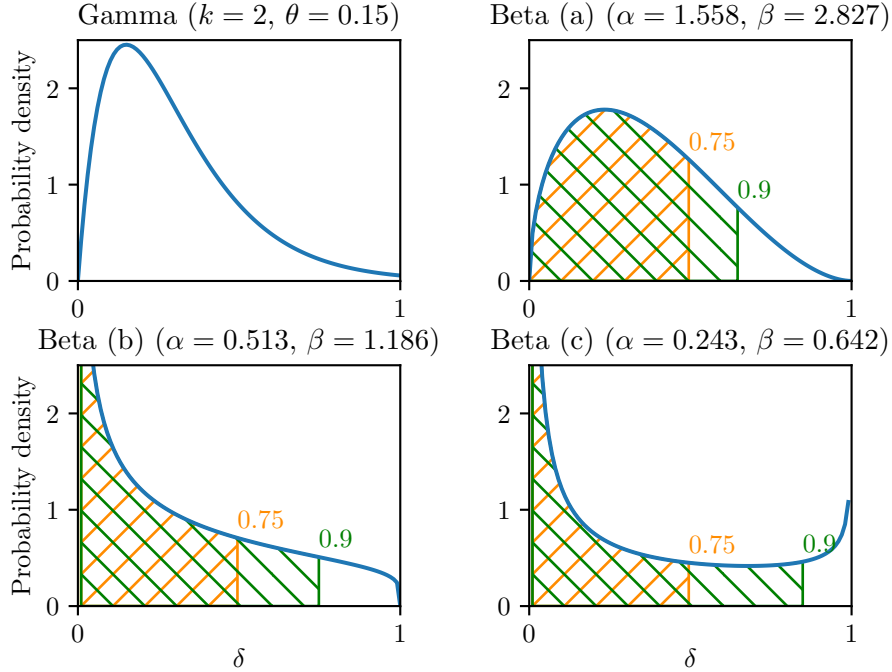}
		\end{center}
		\caption{Shape of the probability density functions for the four different distributions of $\delta$, the original gamma distribution \citep{perez-ortizGraphBasedApproachesOverSampling2015} and the three proposed configurations of the beta distribution (see \cref{sec:ogo-sp})}
		\label{fig:distributions}
	\end{figure}

	\rev{The beta distribution has been applied to model the behaviour of random variables limited to intervals of finite lengths in a wide variety of disciplines. This distribution, in its standard form, is a continuous distribution with probability density function $f(x)$ given by:
	\begin{equation*}
		f(x) = \frac{x^{\alpha-1} (1-x)^{\beta-1}}{B(\alpha, \beta)}
	\end{equation*}
	for $0 < x < 1$ and $\alpha > 0, \beta > 0$, where $B$ is the beta function. Depending on the parameter values, the following properties of the distribution can be outlined:
	\begin{itemize}
		\item If $\alpha > 1$, then $f(0)=0$. Similarly, if $\beta > 1$, then $f(1)=0$.
		\item If both $\alpha > 1$ and $\beta > 1$ (see upper right plot in \cref{fig:distributions}), it has a unique mode at $\frac{\alpha-1}{\alpha + \beta - 2}$.
		\item If $\alpha = \beta$, it is symmetric. If $\alpha = \beta = 1$ it becomes the uniform distribution.
	\end{itemize}
	}
	
	We refer to this new method from now on as \ogosp{} with beta frontier distribution (\ogospbeta{}).
	
	\rev{Based on the two different possibilities of the two endpoints $f(0)$ and $f(1)$, four different asymmetric shapes can be obtained for this distribution. One of these shapes ($\alpha > 1$ and $\beta < 1$) will not be considered, as this would put more probability mass in the neighbouring class side of the distribution. To ensure this, we use the quantile constraint $P(\delta < 0.5) = 0.75$, so that the majority of the probability mass is in the augmented class side.}
		
	\rev{\cite{vandorpSolvingParametersBeta2000} prove that two quantile constraints are sufficient to parametrize the beta distribution and provide a numerical method to obtain the values of $\alpha$ and $\beta$ that satisfy these constraints. We therefore choose three other quantile constraints that, in combination with the previous one, yield the three different shapes, and, using the aforementioned method, we compute the values of $\alpha$ and $\beta$:}
	
	\begin{enumerate}[label=(\alph*)]
		\item Beta distribution where $P(\delta < 0.5) = 0.75$ and $P(\delta < 0.65) = 0.9$: ${\delta \sim \text{Beta}(\alpha=1.558, \beta=2.827)}$
		\item Beta distribution where $P(\delta < 0.5) = 0.75$ and $P(\delta < 0.75) = 0.9$: ${\delta \sim \text{Beta}(\alpha=0.513, \beta=1.186)}$
		\item Beta distribution where $P(\delta < 0.5) = 0.75$ and $P(\delta < 0.85) = 0.9$: ${\delta \sim \text{Beta}(\alpha=0.243, \beta=0.642)}$
	\end{enumerate}

	All three configurations were tested and compared to the original \ogosp{}.

	Our hypothesis is that the beta distribution will be a better candidate for synthetic sample generation for certain datasets, like the one that will be studied in this paper (\cref{sec:data-description}). While configuration (a) imitates the original gamma distribution just for comparison, configuration (b) and (c) of \ogospbeta{} exploit the versatility of the beta distribution.
	
	A graphical visualization of the shape of the probability density function for all distributions can be seen in \cref{fig:distributions}. From there, it can be noted that when $\alpha < 1$ the value of the PDF for $\delta = 0$ tends to infinity, as more probability mass is directed to that extreme. The same thing happens for $\beta < 1$ and $\delta = 1$. This way, the probability of generating samples in the inter-class region can be controlled, while favouring the generation of samples in the augmented class region with respect to the neighbouring class.
	
	The detailed process is finally specified in algorithm \ref{alg:ogo-sp}.
	
	\begin{algorithm}
		\KwIn{$D$: The original dataset}
		\KwIn{$q$: The class to augment}
		\KwIn{$n$: How many instances of class $q$ to generate}
		\KwIn{$k$: The neighbourhood size}
		\Begin(\ogosp{}{($D$, $q$, $n$, $k$)}){
			Construct graph $G_q = (V_q, E_q)$ as described in \cref{sec:ogo-sp} \\
			$D' \leftarrow \emptyset$ \\
			\For{$i=1$ to $n$}{
				\Repeat{$y_i = \class_q \lor y_j = \class_q$}{
					Select a random $e = (\xbf_i, \xbf_j)$ from $E_q$, with uniform probability
				}
				\eIf(class-interior edge case){$y_i = \class_q \land y_j = \class_q$}{
					Sample $\delta$ from an uniform distribution $U(0, 1)$
				}(class-frontier edge case){
					\If(Swap so that $\xbf_i$ is the one with label $\class_q$){$y_i \neq \class_q$}{
						$\xbf_i, \xbf_j \leftarrow \xbf_j, \xbf_i$
					}
					Sample $\delta$ from an asymmetrical distribution (see \cref{sec:ogo-sp})
				}
				$\xbf' \leftarrow (1 - \delta) \xbf_i + \delta \xbf_j$ \\
				$D' \leftarrow D' \cup \{\xbf'\}$
			}
			\Return{$D'$}
		}
		\caption{\ogosp{} algorithm}
		\label{alg:ogo-sp}
	\end{algorithm}
	
	In our case, the data described in \cref{sec:data-description} is severely imbalanced, so data augmentation is crucial for the performance of the classification model. We argue that the \ogospbeta{} algorithm is well suited here due to the ordinal nature of the problem: the alteration of dopaminergic activity is a gradual process, so it is expected to see a more severe damage in a later stage of the disease and vice versa. Moreover, because the intermediate classes are the minority ones, the beta distribution will favour the generation of samples in their regions.
	
	\subsection{Application to spatial data}
		
	Applying techniques like \ac{SMOTE} or \ogosp{} to spatial data, such as images or 3D scans is not appropriate. They are unable to capture the variability of the position of different objects in a scene or anatomical elements on a CT scan. Applying these techniques in the original space that the images are sampled on yields completely unnatural and inappropriate synthetic samples that detract from the generalization capabilities of the resulting model.
	
	On the other hand, the convolutional part of a \ac{CNN} model tries to achieve a projection from the original space to a small number of features that can separate the samples correctly according to the classification problem at hand. The space of this projection should be better suited for interpolation and, consequently, for the application of \ac{SMOTE} and derived techniques. Thus, in this paper, we propose a two-step training process for the application of class balancing:
	
	\begin{enumerate}
		\item First the whole network (convolutional + fully connected parts) is trained on the original dataset $D$.
		\item Once the stopping criterion is reached, the convolutional part $g$ is used to project the original dataset $D$ into a new space with reduced dimensionality in order to obtain $D'$:
		\begin{gather*}
			D' = \{(g(\xbf_i), y_i) \; | \; (\xbf_i, y_i) \in D\}, \\
			g: \mathbb{R}^{d} \rightarrow \mathbb{R}^{d'},
		\end{gather*}
		where $d$ is the original dimensionality of the data and $d'$ is the new reduced dimensionality.
		
		\item \rev{New synthetic samples for each class $q$ are generated by using \ac{SMOTE} in the nominal case and \ogosp{}/\ogospbeta{} in the ordinal case. If $n_q$ is defined as the number of samples labelled as $\class_q$ in $D'$, then $D'_{+q}$ are the generated samples of class $q$. The number of samples to generate for each class is chosen so as to equalize the number of training samples of all classes. Note that this means that no synthetic sample will be generated for the majority class:
		\begin{equation*}
			\left | D'_{+q} \right | = \left ( \max_{1\leq k \leq Q} n_k \right ) - n_q.
		\end{equation*}}
	
		\item \rev{Synthetic samples are then merged with dataset $D'$ to produce the augmented dataset $D'_+$:}
		\begin{equation*}
			\rev{D'_+ = D' \cup \left ( \bigcup_{q=1}^{Q} D'_{+q} \right ).}
		\end{equation*}
	
		\item Finally, only the fully connected part of the original model is trained again using $D'_+$, with the same stopping criterion.
	\end{enumerate}

    \section{Experimentation} \label{sec:experimentation}
    
    The five models previously described (one nominal and four ordinal for the different distributions for $\delta$) will be compared against each other, in order to evaluate the effect of using the ordinal information in the learning process.
   
    \subsection{Experimental design}
    
    \rev{A stratified 5-fold cross-validation over the complete dataset is performed. The dataset is split into 5 (approximately) equal size subsets in a stratified fashion, so that the class distribution is maintained for each fold. For each step, one subset is used for testing and the rest are used as training samples.}
    
    \rev{For each of these 5-fold steps, in the first phase, a model selection process is performed, i.e. the hyperparameters of the algorithm are tuned. For this, three 90/10 holdout splits are performed and all the possible combinations of the following parameter values are considered:
   	\begin{itemize}
   		\item Learning rate ($\eta$): \{ \SIlist[list-pair-separator={, }, retain-unity-mantissa=false]{1e-3;1e-4}{} \}.
   		\item Hidden layer size ($H$): \{2048, 4096\}\footnote{Note that in the case of the ordinal model, each output uses a $(Q-1)$th of this number of nodes for each binary output, so the number of parameters between models is comparable}.
   		\item Convolution kernel size ($k$): \{3, 5\}.
   		\item Neighbourhood size for the data augmentation method: \{3, 5\}.
    \end{itemize}}
    \rev{The mean $\mae$ score across the three splits is used to rank the parameter combinations, and the best combination for each fold is then used for evaluation.}
    
    \rev{Once the hyperparameter selection phase is completed, the optimal parameter combination is used in the second phase for final evaluation. The model is initialized and trained 30 times with different 90/10 train/validation splits, as well as a different random seed for initialization of the weights of the network and data augmentation.}
    
\begin{figure}
    	\includegraphics[width=0.9\linewidth]{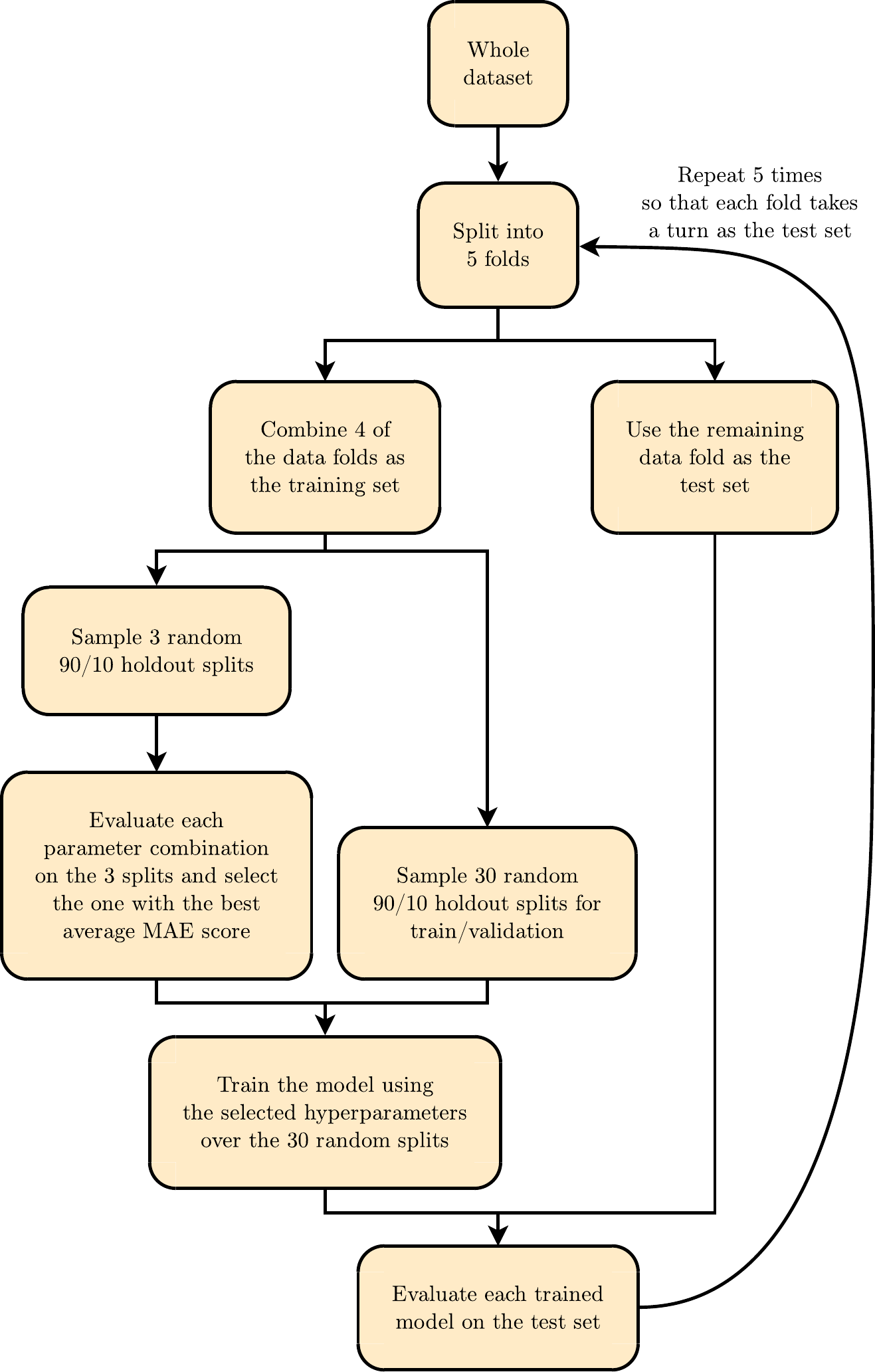}
    	\caption{The cross-validation scheme used for the validation of hyperparameters and evaluation of the models.}
    	\label{fig:cross-validation}
    \end{figure}

    \rev{The same scheme is repeated for each of the data folds. An illustration of this process can be seen in \cref{fig:cross-validation}.}
    
	In every training instance, \rev{the validation samples are used for early stopping of the training process}: when the loss over the validation set does not improve for 50 epochs (\enquote{patience} parameter), training is stopped, and the best performing weights are restored.
	
	A batch size of 32 samples is used in all cases.
	
	The code used to perform the experiments can be accessed through GitHub\footnote{\url{https://github.com/ayrna/ordinal-cnn-parkinsons}}.
	
	\subsection{Evaluation metrics}
	
	\paragraph{Multi-class nominal and ordinal metrics}
	
	\rev{While the \ac{CCR} is usually the most important criterion in classification tasks, in the case of high class imbalance, it is less relevant \citep{provostAnalysisVisualizationClassifier1997}. In scenarios like this, a model that disregards the minority classes can still obtain a high \ac{CCR} score, which is not desirable, and per-class sensitivity needs to be addressed \citep{sanchez-monederoWeightingEfficientAccuracy2011}. Still, while optimizing sensitivity, specificity may suffer a performance hit, so it also needs to be monitored.}
		
	\rev{In addition to this, \ac{CCR} does not consider how much each prediction deviates from the ground truth, as it is designed primarily for nominal classification problems (where all the mistakes are equally penalised). For ordinal classification problems, a classification error of only 1 class is more desirable over an error of 2 classes. For this reason, rank difference metrics like the \ac{MAE}, \ac{SR}, \ac{KR} \citep{cardosoMeasuringPerformanceOrdinal2011} and \ac{WK} \citep{ben-davidComparisonClassificationAccuracy2008} are better suited to evaluate the performance of the model. Moreover, \ac{MAE} may also be deceiving in high class imbalance scenarios, so per-class \ac{MAE} is also useful to consider.}
	
	\rev{When monitoring per-class metrics, it is useful to look at the minimum in order to ensure that performance does not increase at the expense of ignoring some of the classes.}
	
	In this study, the following metrics will be used. Metrics to be maximized are marked with \maximize{} and metrics to be minimized with \minimize. \rev{In the following equations, $N$ denotes the number of test samples, $Q$ denotes the number of classes, $y_i$ is the class label for sample $\xbf_i$ and $\hat{y}_i$ is the estimated label for sample $\xbf_i$.}
	
	\begin{itemize}
		\item \acl{CCR} \maximize{}: $\ccr = \frac{1}{N} \sum_{i=1}^{N} 1\{\hat{y}_i = y_i\}$.
		
		\item Geometric mean of the sensitivities and minimum sensitivity \maximize{}:
		\begin{align*}
			\gms &= \sqrt[Q]{\prod_{q=1}^{Q} S_q}, \\
			\ms &= \min_{1 \leq q \leq Q} S_q,
		\end{align*} where \rev{per-class sensitivity ($S_q$) is}:
		\begin{equation*}
			S_q = \frac{\sum_{i=1}^N 1\{y_i = \hat{y}_i = q\}}{\sum_{i=1}^N 1\{y_i = q\}}.
		\end{equation*}
	
		\item \rev{Geometric mean of the specificities and minimum specificity \maximize{}:}
		\begin{align*}
			\mrev \gmsp & \mrev = \sqrt[Q]{\prod_{q=1}^{Q} Sp_q}, \\
			\mrev \msp & \mrev = \min_{1 \leq q \leq Q} Sp_q,
		\end{align*} \rev{where per-class specificity ($Sp_q$) is:}
		\begin{equation*}
			\rev{Sp_q = \frac{\sum_{i=1}^N 1\{y_i \neq q \land \hat{y}_i \neq q\}}{\sum_{i=1}^N 1\{y_i \neq q\}}.}
		\end{equation*}

		\item \acl{MAE} \minimize{}: $\mae = \frac{1}{N} \sum_{i=1}^{N} \abs{\hat{y}_i - y_i}$.
		
		\item Average and maximum $\mae$ \minimize{}:
		\begin{align*}
			\amae &= \frac{1}{Q} \sum_{q=1}^Q \mae_q, \\
			\mmae &= \max_{1 \leq q \leq Q} \mae_q,
		\end{align*} where \rev{per-class $\mae$ ($\mae_q$) is}:
		\begin{equation*}
			\mae_q = \frac{\sum_{i=1}^N 1\{y_i = q\} \abs{\hat{y}_i - y_i}}{\sum_{i=1}^N 1\{y_i = q\}}.
		\end{equation*}
	
		\item Weighted Cohen's Kappa \maximize{}: \rev{$\wk = 1 - \dfrac{\sum_{i=1}^Q \sum_{j=1}^Q w_{ij} p_{ij}}{\sum_{i=1}^Q \sum_{j=1}^Q w_{ij} e_{ij}}$.}
		
		\rev{where $w_{ij}$ is the disagreement cost when $y = \class_i$ and $\hat{y} = \class_j$ ($w_{ij} = | i - j |$), $p_{ij}$ is the observed agreement and $e_{ij}$ is the expected agreement due to chance.}
		
		\item Kendall rank correlation coefficient \maximize{}: $\kendall = \dfrac{n_c - n_d}{\sqrt{(n_c + n_d + n_1)(n_c + n_d + n_2)}}$,
		
		\rev{where $n_c$, $n_d$, $n_1$ and $n_2$ are computed from every pair $\{(y_i, \hat{y}_i), (y_j, \hat{y}_j)\}$. $n_c$ is the number of concordant pairs, $n_d$ is the number of discordant pairs, and $n_1$ and $n_2$ is the number of tied pairs only for $y$ and only for $\hat{y}$, respectively:}
		\begin{align*}
			\rev{n_c} &\rev{= \sum_{i=1}^N \sum_{j=i+1}^N 1\{[(y_i < y_j) \land (\hat{y}_i < \hat{y}_j)] \lor [(y_i > y_j) \land (\hat{y}_i > \hat{y}_j)]\},} \\
			\rev{n_d} &\rev{= \sum_{i=1}^N \sum_{j=i+1}^N 1\{[(y_i < y_j) \land (\hat{y}_i > \hat{y}_j)] \lor [(y_i > y_j) \land (\hat{y}_i < \hat{y}_j)]\},} \\
			\rev{n_1} &\rev{=  \sum_{i=1}^N \sum_{j=i+1}^N 1\{ (y_i = y_j) \land (\hat{y}_i \neq \hat{y_j}) \},} \\
			\rev{n_2} &\rev{= \sum_{i=1}^N \sum_{j=i+1}^N 1\{ (y_i \neq y_j) \land (\hat{y}_i = \hat{y_j}) \}.}
		\end{align*}
		
		\item Spearman rank correlation coefficient \maximize{}: $\spearman = \dfrac{\Cov(y, \hat{y})}{\sigma_y \sigma_{\hat{y}}}$, \rev{where $\Cov(y, \hat{y})$ is the covariance between the ground truth labels and the predicted labels and $\sigma_y$ and $\sigma_{\hat{y}}$ is their standard deviation.}
	\end{itemize}

	Of the previous metrics, $\ccr$, $\ms$, $\mae$, $\amae$, $\mmae$, $\wk$, $\kendall$ and $\spearman$ are completely defined by  \cite{cruz-ramirezMetricsGuideMultiobjective2014} and $\gms$ by \cite{perez-ortizGraphBasedApproachesOverSampling2015}. $\ccr$, $\gms$, $\ms$, \rev{$\gmsp$ and $\msp$} are all purely nominal metrics, while the rest apply only to ordinal class labels and are generally more relevant for this kind of classification problems.
	
	Also, $\gms$, $\ms$, \rev{$\msp$, $\gmsp$}, $\amae$ and $\mmae$ were chosen as class imbalance-sensitive metrics, to asses the effectiveness in these scenarios.
	
	\rev{Lastly, based on the output scores of the models, the area under the Receiver Operating Characteristic (ROC) curve ($\auc$) will be computed. In order to obtain this value, as many curves as the number of classes are obtained, where each curve is based on a binary one-vs-rest labelling (OVR). Then, the average $\auc$ is obtained from the $Q$ curves.}
	
	\paragraph{Binary metrics}
	
	\rev{Given that no previous literature exists on the diagnosis of the stage of presynaptic damage from \ac{PD}, some binary metrics will be extracted from the results with the only purpose of comparing it to previous works, which deal with the diagnosis problem of discerning \ac{PD} patients from Healthy Controls (HC). Directly comparing $\ccr$ or any other of the previously mentioned metrics against the case of a binary classifier would be incorrect and unfair, as the ordinal metric is quite more demanding, having to classify into four different classes instead of only two.}
	
	\rev{For this purpose, all class 0 labels will be considered as \enquote{negative class} or HC and the rest (class 1 through 3) will be considered as \enquote{positive class} or PD. From this, a confusion matrix can be extracted. Using the number of true positives (TP), true negatives (TN), false positives (FP) and false negatives (FN) the following metrics can be obtained:}
	\begin{align*}
		&\rev{\text{Accuracy} = \frac{\text{TP} + \text{TN}}{\text{TP} + \text{TN} + \text{FP} + \text{FN}},} \\
		&\rev{\text{Sensitivity} = \frac{\text{TP}}{\text{TP} + \text{FN}},} \\
		&\rev{\text{Specificity} = \frac{\text{TN}}{\text{TN} + \text{FP}}.}
	\end{align*}
	
	\subsection{Experimental results}

	\sisetup{detect-weight=true,detect-inline-weight=math}
	\begin{table}
		\caption{Summary of evaluation results. Best mean results are highlighted in bold.}
		\label{tbl:results-summary}
		\begin{adjustbox}{center}
			\small
			\begin{tabular}{@{} l S[table-format=1.4,round-precision=4,round-mode=places] S[table-format=1.4,round-precision=4,round-mode=places] S[table-format=1.4,round-precision=4,round-mode=places] S[table-format=1.4,round-precision=4,round-mode=places] S[table-format=1.4,round-precision=4,round-mode=places] S[table-format=1.4,round-precision=4,round-mode=places] @{}}
\toprule
{} & \multicolumn{2}{c}{{ $\ccr$ \maximize }} & \multicolumn{2}{c}{{ $\gms$ \maximize }} & \multicolumn{2}{c}{{ $\ms$ \maximize }} \\ \cmidrule(lr){2-3} \cmidrule(lr){4-5} \cmidrule(lr){6-7}
{} &                     { Mean } & { Std. dev. } &                     { Mean } & { Std. dev. } &                     { Mean } & { Std. dev. } \\
\midrule
Nominal          & \bfseries 0.7447997152656441 &      0.041189 &           0.1927458189554228 &      0.201516 &          0.06793602693602689 &      0.077447 \\
\ogosp{}         &           0.7121270950624472 &      0.042177 &           0.4256014557281905 &      0.184689 &          0.24256397306397304 &      0.161138 \\
\ogospbeta{} (a) &            0.704823011712936 &      0.048713 &           0.4134423534432952 &      0.189062 &           0.2296734006734005 &      0.157525 \\
\ogospbeta{} (b) &           0.7108444962143278 &      0.064402 &          0.42386896414482494 &      0.173427 &          0.22345959595959594 &      0.147505 \\
\ogospbeta{} (c) &           0.7255464958260535 &      0.042330 & \bfseries 0.4403418299228564 &      0.183046 & \bfseries 0.2459511784511783 &      0.145062 \\
\bottomrule
\end{tabular}

		\end{adjustbox}
		\rev{
		\begin{adjustbox}{center}
			\small
			\begin{tabular}{@{} l S[table-format=1.4,round-precision=4,round-mode=places] S[table-format=1.4,round-precision=4,round-mode=places] S[table-format=1.4,round-precision=4,round-mode=places] S[table-format=1.4,round-precision=4,round-mode=places] S[table-format=1.4,round-precision=4,round-mode=places] S[table-format=1.4,round-precision=4,round-mode=places] @{}}
\toprule
{} & \multicolumn{2}{c}{{ $\msp$ \maximize }} & \multicolumn{2}{c}{{ $\gmsp$ \maximize }} & \multicolumn{2}{c}{{ $\auc$ \maximize }} \\ \cmidrule(lr){2-3} \cmidrule(lr){4-5} \cmidrule(lr){6-7}
{} &                     { Mean } & { Std. dev. } &                     { Mean } & { Std. dev. } &                    { Mean } & { Std. dev. } \\
\midrule
Nominal          &           0.7691248244273924 &      0.056865 &           0.8898325156838611 &      0.017879 &          0.8486356604124708 &      0.034364 \\
\ogosp{}         &           0.8276138809063003 &      0.040747 &           0.8979133085312103 &      0.013839 &          0.8588045398259793 &      0.034470 \\
\ogospbeta{} (a) &            0.822280291837163 &      0.040584 &           0.8956604258531666 &      0.014647 &          0.8552784674260819 &      0.038515 \\
\ogospbeta{} (b) &           0.8293755806292908 &      0.060139 &           0.8976947880759131 &      0.022935 &          0.8566651826180915 &      0.033127 \\
\ogospbeta{} (c) & \bfseries 0.8311799074927395 &      0.051152 & \bfseries 0.9008663106399062 &      0.017030 & \bfseries 0.859552554333348 &      0.036572 \\
\bottomrule
\end{tabular}

		\end{adjustbox}
		}
		\begin{adjustbox}{center}
			\small
			\begin{tabular}{@{} l S[table-format=1.4,round-precision=4,round-mode=places] S[table-format=1.4,round-precision=4,round-mode=places] S[table-format=1.4,round-precision=4,round-mode=places] S[table-format=1.4,round-precision=4,round-mode=places] S[table-format=1.4,round-precision=4,round-mode=places] S[table-format=1.4,round-precision=4,round-mode=places] @{}}
\toprule
{} & \multicolumn{2}{c}{{ $\mae$ \minimize }} & \multicolumn{2}{c}{{ $\amae$ \minimize }} & \multicolumn{2}{c}{{ $\mmae$ \minimize }} \\ \cmidrule(lr){2-3} \cmidrule(lr){4-5} \cmidrule(lr){6-7}
{} &                     { Mean } & { Std. dev. } &                     { Mean } & { Std. dev. } &                     { Mean } & { Std. dev. } \\
\midrule
Nominal          &          0.38262796867922066 &      0.073664 &           0.6802538072739684 &      0.112743 &           1.1427323232323237 &      0.132811 \\
\ogosp{}         &          0.37381738173817347 &      0.058617 &           0.5670550260671225 &      0.113055 &           0.9043164983164983 &      0.214830 \\
\ogospbeta{} (a) &          0.37908496732026126 &      0.064325 &           0.5668393516579003 &      0.107002 & \bfseries 0.9020656565656564 &      0.200475 \\
\ogospbeta{} (b) &           0.3729398822235161 &      0.075076 &            0.563056810423746 &      0.103824 &           0.9187118807118805 &      0.180004 \\
\ogospbeta{} (c) & \bfseries 0.3638814469682262 &      0.064949 & \bfseries 0.5594134024577573 &      0.118579 &           0.9112996632996632 &      0.189582 \\
\bottomrule
\end{tabular}

		\end{adjustbox}
		\begin{adjustbox}{center}
			\small
			\begin{tabular}{@{} l S[table-format=1.4,round-precision=4,round-mode=places] S[table-format=1.4,round-precision=4,round-mode=places] S[table-format=1.4,round-precision=4,round-mode=places] S[table-format=1.4,round-precision=4,round-mode=places] S[table-format=1.4,round-precision=4,round-mode=places] S[table-format=1.4,round-precision=4,round-mode=places] @{}}
\toprule
{} & \multicolumn{2}{c}{{ $\tau_b$ \maximize }} & \multicolumn{2}{c}{{ $\kappa$ \maximize }} & \multicolumn{2}{c}{{ $\spearman$ \maximize }} \\ \cmidrule(lr){2-3} \cmidrule(lr){4-5} \cmidrule(lr){6-7}
{} &                     { Mean } & { Std. dev. } &                     { Mean } & { Std. dev. } &                     { Mean } & { Std. dev. } \\
\midrule
Nominal          &            0.711896578433911 &      0.056489 &           0.6815750633390985 &      0.064491 &           0.7701889750817208 &      0.059990 \\
\ogosp{}         &           0.7322767532478356 &      0.044160 &           0.6822011726718525 &      0.054792 &           0.7980282436952416 &      0.051622 \\
\ogospbeta{} (a) &           0.7281771321178986 &      0.049595 &            0.677644402841616 &      0.057951 &           0.7967652672769455 &      0.052670 \\
\ogospbeta{} (b) &           0.7342571623024985 &      0.046746 &           0.6844046117222574 &      0.065353 & \bfseries 0.8010916993042552 &      0.051182 \\
\ogospbeta{} (c) & \bfseries 0.7388654350175323 &      0.048095 & \bfseries 0.6897646191355054 &      0.059587 &           0.7986035683661348 &      0.054751 \\
\bottomrule
\end{tabular}

		\end{adjustbox}
	\end{table}

	\begin{figure}
		\begin{adjustbox}{center}
			\input{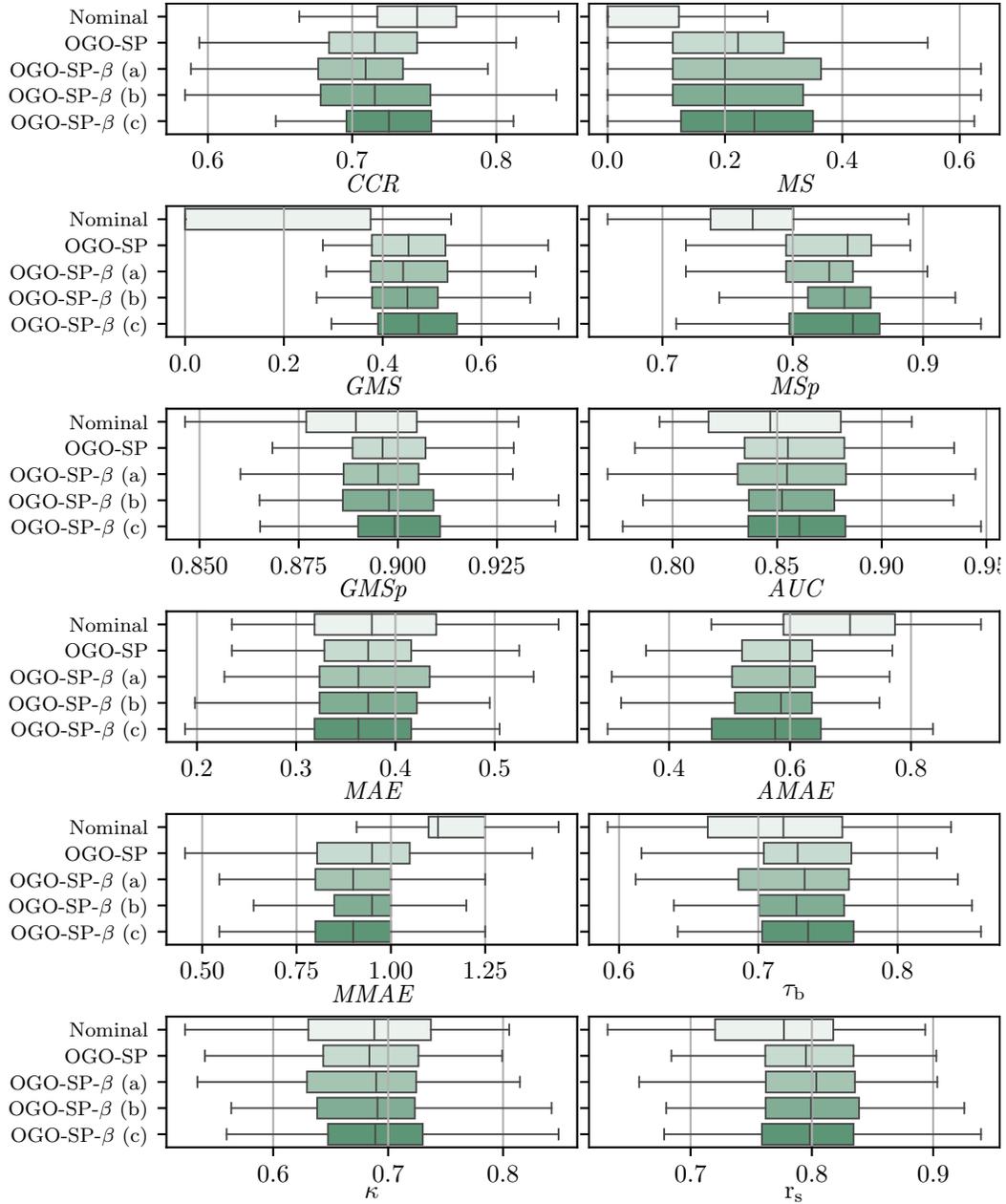}
		\end{adjustbox}
		\rev{
		\caption{Graphical summary of the experimentation results as boxplots.}
		\label{fig:results-summary}
		}
	\end{figure}

	\begin{figure}
		\begin{adjustbox}{center}
			\input{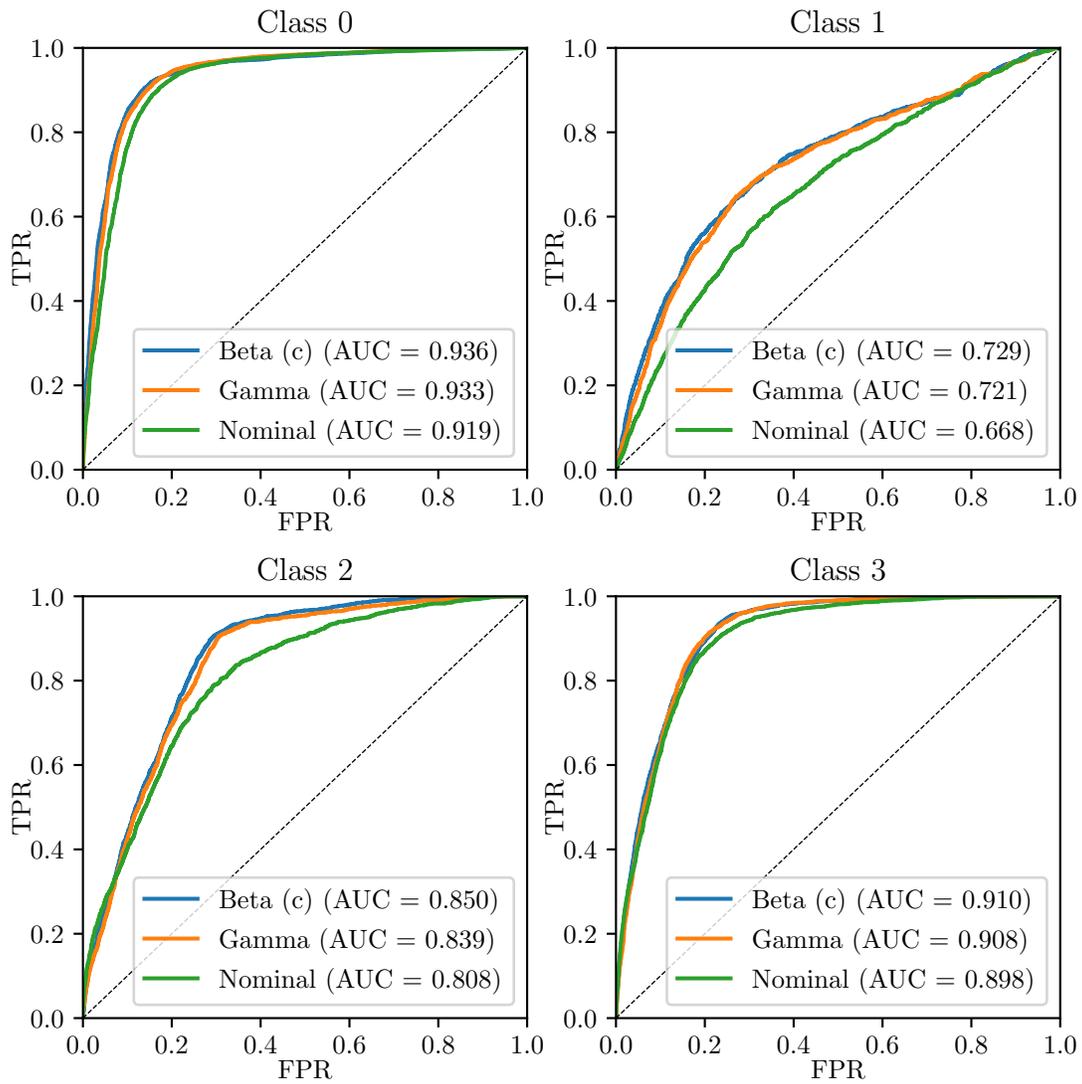}
		\end{adjustbox}
		\rev{\caption{ROC obtained for each of the four classes by three of the evaluated methodologies. The curves are obtained according to the output scores of each model using a one-vs-rest labelling (OVR) for each class.}
		\label{fig:aucs}}
	\end{figure}

	\Cref{tbl:results-summary} includes a summary of all the results from the \rev{150} executions of the different methodologies, based on the different performance metrics introduced in the previous subsection. A graphical summary of the results can be found in \cref{fig:results-summary}.
	
	\rev{From these results, it can be noted that some configuration of \ogospbeta{} always shows better average performance over all other classifiers in all metrics, with the exception of $\ccr$ for which the nominal methodology obtains the best results, with \ogospbeta{} (c) in second place. By looking at the $\gms$ and $\ms$ metrics, it is very clear that the nominal methodology fails to address the present class imbalance and ignores the minority classes, arranging all test samples into a couple of the majority classes, which drops the score to zero in most of the evaluation splits.}
	
	\rev{Comparing \ogospbeta{} to the original \ogosp{}, \ogospbeta{} generally obtains better average performance over all splits, specially according to $\mae$.}
	
	\rev{Both ordinal methodologies clearly outperform the nominal one. \cref{fig:aucs} shows the corresponding ROCs for the different classes of the problem. Both ordinal methodologies obtain a noticeable advantage in the intermediate minority classes, specially for class 1, the class with the least number of samples.}
	
	We used a Wilcoxon signed-rank test to detect significant differences between the metrics. This version of the test was used to measure the performance of the classifiers, because the variables come from the same sample \citep{wilcoxonIndividualComparisonsRanking1945}. Purely nominal methodology was compared with the overall best performing ordinal methodology using \ogospbeta{} (configuration (c)). We also used the test to compare the ordinal methodology using \ogosp{} with the best performing ordinal methodology using \ogospbeta{} (configuration (c)).
	
	The critical values ($p$-values), for a bilateral test of each metric, are provided in \cref{tbl:wilcoxon}.
	
	\sisetup{detect-all}
	\begin{table}
		\caption{$p$-values for the two-tailed Wilcoxon signed rank test. \enquote{N} stands for the nominal methodology, \enquote{$\gamma$} stands for the methodology using the original \ogosp{} algorithm and \enquote{$\beta$} stands for the methodology using the \ogospbeta{} algorithm. Values less than $\alpha = 0.05$ have been highlighted in bold. For these tests, the (+) subscript denotes an advantage for \ogospbeta{} and (--) otherwise.}
		\label{tbl:wilcoxon}
		
		\begin{center}
			\small
			\setlength{\tabcolsep}{3pt}
			\begin{tabular}{@{} l S[table-column-width=70pt,table-space-text-post={\subscript{(+)}},table-format = 1.1e+2] S[table-column-width=70pt,table-space-text-post={\subscript{(+)}},table-format = 1.1e+2] S[table-column-width=70pt,table-space-text-post={\subscript{(+)}},table-format = 1.1e+2]}
\toprule
{} &                           { $\ccr$ } &                          { $\gms$ } &                           { $\ms$ } \\
\midrule
N vs. $\beta$ (c)         & \bfseries 3.5e-05 {\subscript{(--)}} & \bfseries 9.6e-18 {\subscript{(+)}} & \bfseries 6.4e-19 {\subscript{(+)}} \\
$\gamma$  vs. $\beta$ (c) &  \bfseries 9.5e-04 {\subscript{(+)}} &                             2.7e-01 &                             7.6e-01 \\
\bottomrule
\end{tabular}
 \\[1em]
			
			\rev{
			\begin{tabular}{@{} l S[table-column-width=70pt,table-space-text-post={\subscript{(+)}},table-format = 1.1e+2] S[table-column-width=70pt,table-space-text-post={\subscript{(+)}},table-format = 1.1e+2] S[table-column-width=70pt,table-space-text-post={\subscript{(+)}},table-format = 1.1e+2]}
\toprule
{} &                          { $\msp$ } &                         { $\gmsp$ } &                          { $\auc$ } \\
\midrule
N vs. $\beta$ (c)         & \bfseries 3.0e-17 {\subscript{(+)}} & \bfseries 2.4e-12 {\subscript{(+)}} & \bfseries 7.7e-06 {\subscript{(+)}} \\
$\gamma$  vs. $\beta$ (c) &                             2.5e-01 & \bfseries 1.9e-02 {\subscript{(+)}} &                             3.9e-01 \\
\bottomrule
\end{tabular}
 \\[1em]
			}
			
			\begin{tabular}{@{} l S[table-column-width=70pt,table-space-text-post={\subscript{(+)}},table-format = 1.1e+2] S[table-column-width=70pt,table-space-text-post={\subscript{(+)}},table-format = 1.1e+2] S[table-column-width=70pt,table-space-text-post={\subscript{(+)}},table-format = 1.1e+2]}
\toprule
{} &                          { $\mae$ } &                         { $\amae$ } &                         { $\mmae$ } \\
\midrule
N vs. $\beta$ (c)         & \bfseries 7.5e-04 {\subscript{(+)}} & \bfseries 5.1e-23 {\subscript{(+)}} & \bfseries 1.3e-18 {\subscript{(+)}} \\
$\gamma$  vs. $\beta$ (c) & \bfseries 3.8e-02 {\subscript{(+)}} &                             1.7e-01 &                             7.3e-01 \\
\bottomrule
\end{tabular}
 \\[1em]
			
			\begin{tabular}{@{} l S[table-column-width=70pt,table-space-text-post={\subscript{(+)}},table-format = 1.1e+2] S[table-column-width=70pt,table-space-text-post={\subscript{(+)}},table-format = 1.1e+2] S[table-column-width=70pt,table-space-text-post={\subscript{(+)}},table-format = 1.1e+2]}
\toprule
{} &                      { $\kendall$ } & { $\kappa$ } &                     { $\spearman$ } \\
\midrule
N vs. $\beta$ (c)         & \bfseries 9.2e-09 {\subscript{(+)}} &      9.3e-02 & \bfseries 6.8e-10 {\subscript{(+)}} \\
$\gamma$  vs. $\beta$ (c) &                             7.3e-02 &      8.0e-02 &                             6.7e-01 \\
\bottomrule
\end{tabular}

		\end{center}
	\end{table}

	\rev{Significant differences in performance (with $\alpha = 0.05$) favouring the ordinal methodology over the nominal one are found for all metrics, but $\ccr$ and $\kappa$. As expected, the nominal methodology fixates on improving $\ccr$ instead of the rest of ordinal metrics, which are generally more desirable for ordinal type classification problems. Moreover, the nominal methodology tends to ignore minority classes, obtaining significantly worse results for $\gms$, $\ms$, $\msp$, $\gmsp$ and $\auc$. Furthermore, \ogospbeta{} performs significantly better in $\ccr$, $\mae$ and $\gmsp$ compared to the original \ogosp{} using the gamma distribution.}
	
	\rev{Clearly, even when the tests do not find significant differences, \ogospbeta{} shows better overall performance in the ordinal metrics. The inconclusiveness present in the imbalance-sensitive metrics ($\gms$, $\ms$, $\msp$ and $\mmae$) could be explained by their unstable nature, which results in a larger standard deviation, and thus it is more difficult for the test to draw a conclusion.}
	
	\rev{Lastly, the results for the binary metrics results can be found in \cref{tbl:binary}. The results are compared against the following works:
		\begin{itemize}
			\item \cite{rizzoAccuracyClinicalDiagnosis2016}: A meta-analysis of 20 different studies, all using different techniques, between 1988 and 2014.
			\item \cite{fuente-fernandezRoleDaTSCANClinical2012}: An aggregation of 2 different studies both using SPECT imaging.
			\item \cite{martinez-murcia3DConvolutionalNeural2017}: A CNN approach for binary classification from SPECT imaging.
			\item \cite{orozco-arroyaveAutomaticDetectionParkinson2016}: An application of radial base Support Vector Machines to running speech audio samples from patients.
			\item \cite{elmaachiDeep1DConvnetAccurate2020}: An application of neural networks to gait sensor data from patients.
		\end{itemize}}
	
	\rev{Note that the task considered in these works is different, as the different possibilities for positive labels (PD) are not differentiated, greatly reducing the complexity. However, even considering that the proposed models are more informative and taking into account that the experimental settings and the datasets are not same, we can conclude that the performance obtained by the different proposals is competitive, specially when trying to achieve a balance between the three binary metrics. That is, the extra information provided by the proposed multi-class classifiers is not obtained at the cost of losing performance in the binary task.}
	
	\sisetup{range-phrase=--}
	\renewcommand*\footnoterule{}
	\begin{table}[]
		\rev{\caption{Binary metrics results of the five proposed methodologies along with three other works. Rows marked as type \enquote{A} correspond to an automatic method while rows marked as type \enquote{H} correspond to studies about diagnosis by human experts. The Data column indicates the reference data used for diagnosis: imaging (I), speech (S), gait (G) or a refined clinical diagnosis (C).}
		\label{tbl:binary}}
		\begin{minipage}{\linewidth}
			\rev{
			\setlength{\tabcolsep}{3pt}
			\begin{adjustbox}{center}
			\begin{tabularx}{1.07\textwidth}{lccccc}
				\toprule
				Method/work                                                                                                                        & Type & Data &          Accuracy          &         Sensitivity          & Specificity                  \\ \midrule
				Nominal                                                                                                                            &  A   &  I   &    \SI{87.62}{\percent}    &     \SI{77.52}{\percent}     & \SI{93.84}{\percent}         \\
				OGO-SP                                                                                                                             &  A   &  I   &    \SI{87.30}{\percent}    &     \SI{86.74}{\percent}     & \SI{87.64}{\percent}         \\
				OGO-SP-B (a)                                                                                                                       &  A   &  I   &    \SI{86.94}{\percent}    &     \SI{86.67}{\percent}     & \SI{87.09}{\percent}         \\
				OGO-SP-B (b)                                                                                                                       &  A   &  I   &    \SI{87.14}{\percent}    &     \SI{87.10}{\percent}     & \SI{87.16}{\percent}         \\
				OGO-SP-B (c)                                                                                                                       &  A   &  I   &    \SI{88.19}{\percent}    &     \SI{86.21}{\percent}     & \SI{89.38}{\percent}         \\
				Rizzo et al. (2016)    \footnote{Values are taken from the refined clinical diagnosis by experts.}                                 &  H   &  C   &    \SI{86.4}{\percent}     &     \SI{86.6}{\percent}      & \SI{84.7}{\percent}          \\
				de la Fuente-Fernández (2012)  \footnote{Ranges represent the performance between early and established diagnosis, when provided.} &  H   &  I   & \SIrange{84}{98}{\percent} &      \SI{98}{\percent}       & \SIrange{67}{94}{\percent}   \\
				Martinez-Murcia et al. (2017)                                                                                                      &  A   &  I   &    \SI{95.5}{\percent}     &     \SI{96.2}{\percent}      & -                            \\ 
				Orozco-Arroyave et al. (2016) \footnote{Ranges represent the worst and best performance across all three tested languages.}        &  A   &  S   & \SIrange{85}{97}{\percent} & \SIrange{76.7}{98}{\percent} & \SIrange{93.3}{96}{\percent} \\
				El Maachi et al. (2020)                                                                                                            &  A   &  G   &    \SI{98.7}{\percent}     &     \SI{98.1}{\percent}      & \SI{100}{\percent}           \\ \bottomrule
			\end{tabularx}
			\end{adjustbox}}
		\end{minipage}
	\end{table}

    \section{Conclusions} \label{sec:conclusions}
    
    We have confirmed experimentally that the exploitation of ordinal information can improve the performance of a complex task such as the assessment of brain activity alteration in \ac{PD}. This exploitation comes from aspects such as the model architecture, the optimization target and the data augmentation strategy. This expands on the current range of models capable of exploiting ordinal information, in this case, a fully 3D \ac{CNN}.
    
    This methodology is able to alleviate the class imbalance problem while improving ordinal performance metrics. This approach could be applied to already existing ordinal classification tasks which suffer from this same problem, which is common among the medical field.
    
    \rev{Furthermore, the proposed \ogospbeta{} ordinal augmentation algorithm improves performance of ordinal and nominal metrics compared to the original \ogosp{}.}
    
    \rev{From the more classical binary diagnosis point of view, a good performance is achieved, as can be noted in the comparison against expert diagnoses and other ML techniques.}
    
    Future work lines include further data acquisition or the application of this methodology to publicly available data. Current available data commonly lacks ordinal information, but ordinal target labels may be extracted from already available information. Larger datasets could help obtain more relevant and precise results. Also, 3D ordinal applications outside the medical can be explored to asses the performance over nominal approaches.
    
    \bibliographystyle{model5-names}
    \bibliography{bibliography}
	
\end{document}